\documentclass[10pt,twocolumn,letterpaper]{article}

\usepackage{iccv}
\usepackage{times}
\usepackage{epsfig}
\usepackage{graphicx}
\usepackage{amsmath}
\usepackage{amssymb}
\usepackage{algorithm}
\usepackage{algorithmic}
\usepackage{setspace}
\usepackage{xcolor}
\usepackage{booktabs}
\usepackage{makecell}
\usepackage{multirow}
\usepackage{float}
\usepackage{marvosym}

\newcommand{\headrule}{\toprule}
\newcommand{\normrule}{\midrule}

\floatname{algorithm}{Procedure}

\usepackage[pagebackref=true,breaklinks=true,colorlinks,bookmarks=false]{hyperref}

\newcommand\blfootnote[1]{%	
  \begingroup
  \renewcommand\thefootnote{}\footnote{#1}%
  \addtocounter{footnote}{-1}%
  \endgroup
}

\iccvfinalcopy

\def\iccvPaperID{7800} % *** Enter the ICCV Paper ID here

\ificcvfinal\fi

\def\GREY{\leavevmode\rlap{\hbox to \hsize{\color{grey!10}\leaders\hrule height .8\baselineskip depth .5ex\hfill}}}
\definecolor{mygray}{gray}{0.4}

\newcommand{\algomusu}{
\begin{algorithm}[t]
\setstretch{1.0}
\caption{\textbf{: Mu}tual \textbf{Su}pervision algorithm}\label{algo}
\begin{algorithmic}[1]
\small
\REQUIRE $\mathcal{G}$, $\mathcal{A}$
\newline $\mathcal{G}$ is a set of ground-truth boxes
\newline $\mathcal{A}$ is a set of all anchors across all pyramid levels
\ENSURE $\mathcal{L}^{\mathit{cls}}, \mathcal{L}^{\mathit{reg}}$
\newline $\mathcal{L}^{\mathit{cls}}$ is the total loss for the classification head
\newline $\mathcal{L}^{\mathit{reg}}$ is the total loss for the regression head
\STATE initialize $w^{\mathit{cls}}_{i}$ and $w^{\mathit{reg}}_{i}$ to zeros for all $i\in \mathcal{A}$;
\FOR{every ground-truth box $j\in \mathcal{G}$}
    \STATE \textcolor{mygray}{\footnotesize // construct candidate bag $\mathcal{C}_j$ for box $j$}
    \STATE $\mathcal{A}_j \leftarrow$ \{$i \in\mathcal{A}: $ anchor $i$ lies in the box $j$ and has maximum IoU with $j$ across $\mathcal{G}$\}; 
    \STATE compute $P_i$ in Equ~\ref{equ:joint} for every anchor $i \in \mathcal{A}_j$;
    \STATE $\mathcal{C}_j \leftarrow \{i\in \mathcal{A}_j: P_i> b\cdot \max_{k\in \mathcal{A}_j} P_k\}$; 
    
    \STATE \textcolor{mygray}{\footnotesize // compute candidate rankings in candidate bags $\mathcal{C}_j$}
    \STATE compute $v^{\mathit{cls}}_{i}$, $v^{\mathit{reg}}_{i}$ for $i\in \mathcal{C}_j$ according to Equ~\ref{equ:value};
    \STATE sort $v^{\mathit{cls}}_{i}$ and $v^{\mathit{reg}}_{i}$ in descending order and obtain ranking $R^{\mathit{cls}}_i$ and $R^{\mathit{reg}}_i$ (starting from $0$);
    
    \STATE \textcolor{mygray}{\footnotesize // compute base losses and transform rankings to weights}
    \STATE assign box $j$ to $i$ in $\mathcal{C}_j$ as detection target and compute base losses for all $i$ in $\mathcal{C}_j$;
    \STATE translate $R^{\mathit{cls}}_i$, $R^{\mathit{reg}}_i$ to $w^{\mathit{cls}}_i$ and $w^{\mathit{reg}}_i$ according to Equ~\ref{equ:translate};
\ENDFOR
\STATE caculate $\mathcal{L}^{\mathit{cls}}$ and $\mathcal{L}^{\mathit{reg}}$ according to Equ~\ref{equ:total_loss};
  \end{algorithmic}
\end{algorithm}
}

\newcommand{\figA}{
\begin{figure}[t]
\centering
\includegraphics[width=\columnwidth]{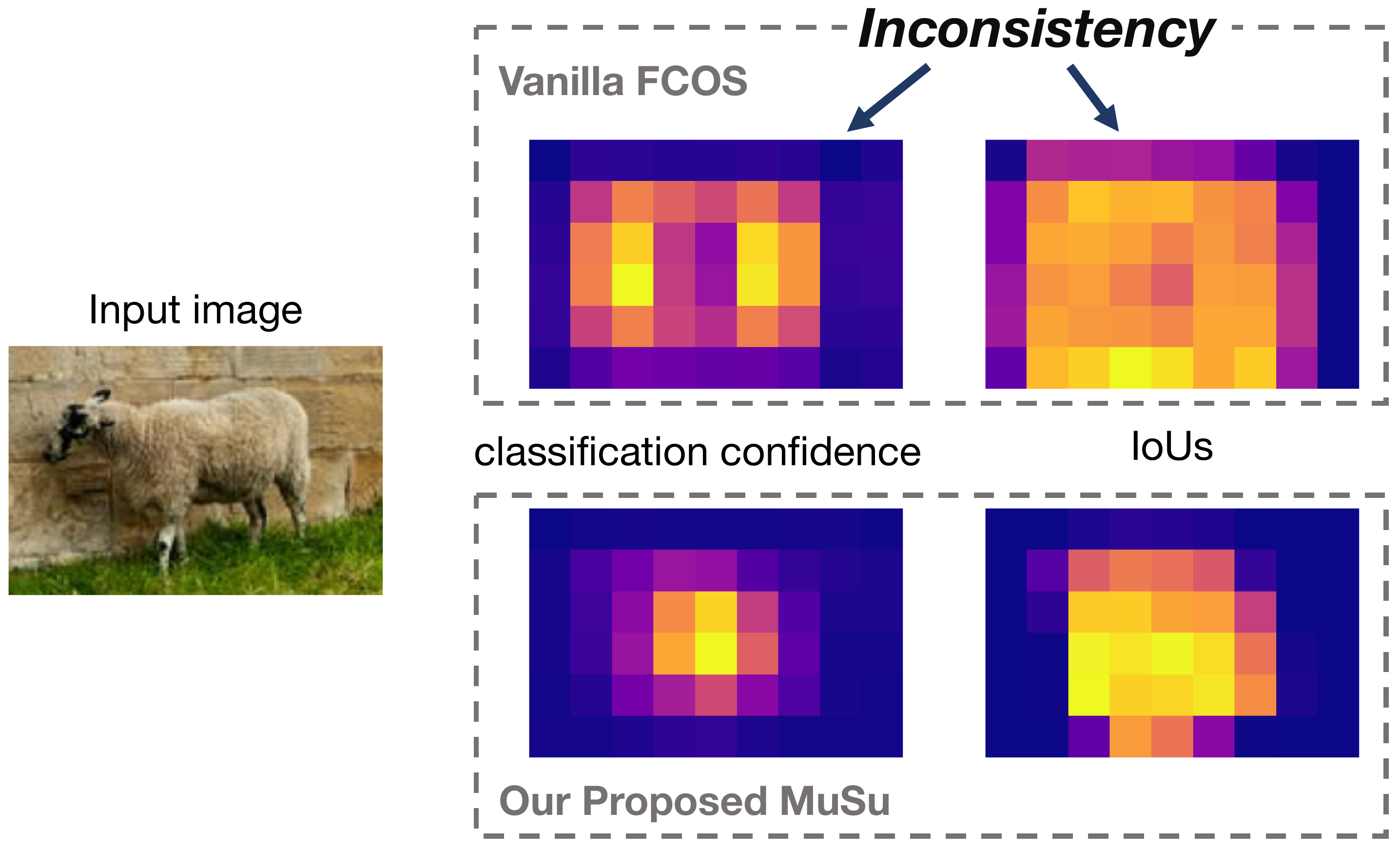}
\caption{
\textbf{Inconsistency from the classification head and regression head} between spatial distributions of classification confidence and IoUs with ground truth predicted in a converged FCOS detector and our MuSu-trained detector. The brighter the pixel looks, the higher the value stands for. The classification confidence is a product of the output of the classification head and centerness estimation as FCOS does in the NMS process. Note that this input image is a training image in MS COCO and the converged FCOS still suffers the inconsistency between classification and regression head. Our MuSu alleviates this inconsistency. \vspace{-0.8em}}
\label{fig:inconsistency}
\end{figure}
}

\newcommand{\figB}{
\begin{figure*}[t]
\centering
\includegraphics[width=0.95\textwidth, trim={0em 4cm 0em 2cm},clip]{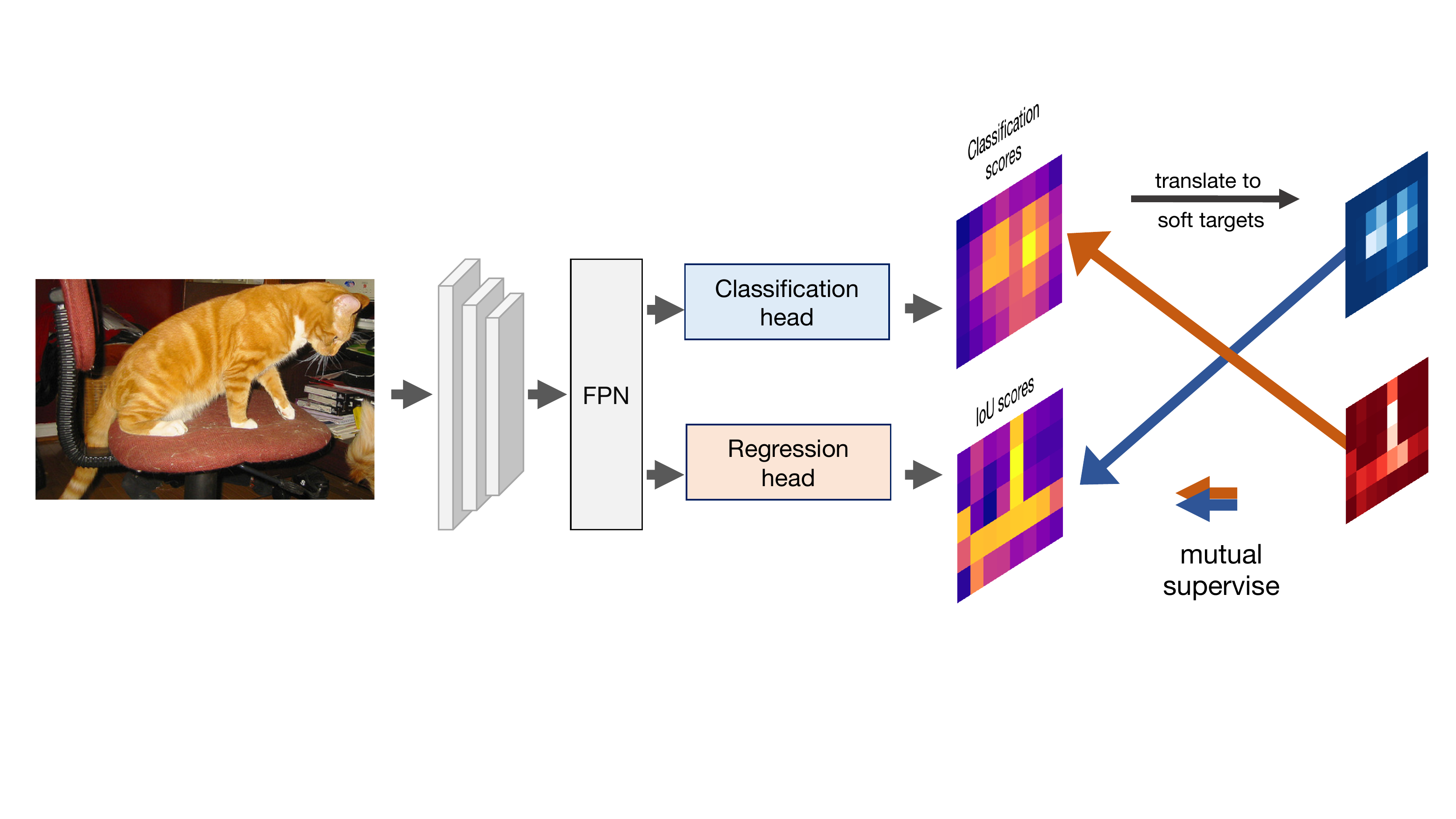}
\caption{
\textbf{Illustration of the mutual supervision (MuSu)} for the classification and regression head. We apply soft targets to supervise by weighting base loss at each location for these two heads respectively. The weights for the regression head are mainly decided by the classification score and the weights for the classification head are in turn mostly based on the localization scores (IoU scores). For more clarity, we do not show the construction of candidate bags and multi-level predictions. \vspace{-0.0em}
}
\label{fig:method}
\end{figure*}
}

\newcommand{\tabA}{
\begin{table}[t]
    \small
    \centering
    \setlength{\tabcolsep}{4pt}
    \begin{tabular}{l|c|cc|ccc}
    \makecell[c]{method}   & AP & AP$_{50}$ & AP$_{75}$ & AP$_{S}$ & AP$_{M}$ & AP$_{L}$ \\
    \headrule
    FCOS               & 36.5 & 55.7 & 38.3 & 21.2 & 40.3 & 48.1 \\
    FCOS++             & 38.6 & 57.2 & 41.7 & 22.4 & 42.4 & 50.2 \\
    \normrule
        \multicolumn{7}{l}{{MuSu with }} \\
    \hspace{0.2em} $v^{\mathit{cls}}_i=v^{\mathit{reg}}_i=p_i$
                       & 38.3 & \textbf{60.0} & 40.5 & 22.8 & 41.5 & 49.3 \\
    \hspace{0.2em} $v^{\mathit{cls}}_i=v^{\mathit{reg}}_i=q_i$
                       & 31.8 & 49.9 & 33.7 & 15.0 & 34.9 & 45.2 \\
    \normrule
    \multicolumn{7}{l}{{MuSu under $v^{\mathit{cls}}_i$ and $v^{\mathit{reg}}_i$ in Equ~\ref{equ:value} with }} \\
    \hspace{0.2em} $\alpha=1.0$       & 40.4 & 59.6 & 43.9 & \textbf{23.5} & 43.5 & 53.7 \\
    \hspace{0.2em} $\alpha=1/2$       & 40.3 & 59.1 & 43.8 & 23.1 & 43.7 & 53.5 \\
    \hspace{0.2em} $\alpha=1/3$       & \textbf{40.6} & 58.9 & \textbf{44.3} & 23.0 & 44.0 & \textbf{54.2} \\
    \hspace{0.2em} $\alpha=1/4$       & 40.5 & 58.9 & 43.8 & 23.4 & \textbf{44.2} & 53.6 \\
    \hspace{0.2em} $\alpha=1/6$       & 40.4 & 59.0 & 43.6 & 22.5 & \textbf{44.2} & 53.9 \\
    \hspace{0.2em} $\alpha=0$       & 38.5 & 57.5 & 41.4 & 20.9 & 42.9 & 52.4 \\
    \end{tabular}
    \vspace{0.3em}
    \caption{\textbf{MuSu criteria value settings for each head} on COCO \texttt{minival} set with ResNet-50 backbone (same in tables below, unless otherwise specified). FCOS and FCOS++ are baselines. FCOS++ denotes the improved FCOS with tricks (e.g., center sampling). The term $\alpha$ refers to the regularizing factor in Equation~\ref{equ:value}. The settings $v^{(\cdot)}=p$ and $v^{(\cdot)}=q$ stand for that the method assigns the same weights for each head, totally according to the classification or localization score of an anchor inside the bag.\vspace{-1em}}
    \label{tab:effective}
\end{table}
}

\newcommand{\tabB}{
\begin{table}[t]
    \footnotesize
    \centering
    \renewcommand\arraystretch{1.0}
    \setlength{\tabcolsep}{6.2pt}
    \begin{tabular}{c|c|cc|ccc}
    $b$
                 & AP & AP$_{50}$ & AP$_{75}$ & AP$_{S}$ & AP$_{M}$ & AP$_{L}$ \\
    \headrule
    $0.20$       & 40.3 & 58.8 & 43.7 & 22.8 & 43.4 & 53.4 \\
    $0.10$       & \textbf{40.6} & \textbf{58.9} & \textbf{44.3} & 23.0 & \textbf{44.0} & \textbf{54.2} \\
    $0.05$       & 40.4 & \textbf{58.9} & 43.7 & \textbf{23.5} & 43.8 & 54.0 \\
    \end{tabular}
    \vspace{0.3em}
    \caption{\textbf{Different threshold coefficients $b$} in Equation~\ref{equ:threshold}.\vspace{-1.0em}}
    \label{tab:anchor_bag}
\end{table}
}

\newcommand{\tabC}{
\begin{table}[t]
    \footnotesize
    \centering
    \renewcommand\arraystretch{1.0}
    \setlength{\tabcolsep}{4pt}
    \begin{tabular}{c|c|cc|ccc}
    \makecell[c]{$(\tau^{\mathit{cls}}$, $\tau^{\mathit{reg}})$ }
                       & AP & AP$_{50}$ & AP$_{75}$ & AP$_{S}$ & AP$_{M}$ & AP$_{L}$ \\
    \headrule
    $(10.0, 5.0)$      & 40.1 & 58.5 & 43.3 & 22.2 & 43.4 & 53.7 \\
    $(5.0, 2.5)$       & 39.9 & 58.5 & 43.4 & 22.1 & 43.5 & 53.2 \\
    \normrule
    $(\sqrt{|\mathcal{C}_j|}, 0.5\sqrt{|\mathcal{C}_j|})$       & \textbf{40.6} & \textbf{58.9} & \textbf{44.3} & \textbf{23.0} & \textbf{44.0} & \textbf{54.2} \\
    \end{tabular}
    \vspace{0.3em}
    \caption{\textbf{Adaptive temperatures $\tau^{\mathit{cls}}$ and $\tau^{\mathit{reg}}$ benefit}. First two rows act as counterparts with fixed temperatures.\vspace{-0.7em}}
    \label{tab:temperature}
\end{table}
}

\newcommand{\tabD}{
\begin{table}[b]
    \footnotesize
    \centering
    \renewcommand\arraystretch{1.0}
    \setlength{\tabcolsep}{8pt}
    \begin{tabular}{c|ccc}
    \makecell[c]{$\tau^{\mathit{cls}}$: $\tau^{\mathit{reg}}$ }   & AP & AP$_{50}$ & AP$_{75}$ \\
    \headrule
    $1:1$       & 40.4 & 58.6 & 44.0 \\
    $1.5:1$     & \textbf{40.6} & 58.9 & 44.1 \\
    $2:1$       & \textbf{40.6} & 58.9 & \textbf{44.3} \\
    $3:1$       & 40.2 & \textbf{59.0} & 43.7 \\
    \end{tabular}
    \vspace{0.3em}
    \caption{\textbf{Varying the ratio of temperature coefficients} of the classification to the regression $\tau^{\mathit{cls}}$: $\tau^{\mathit{reg}}$.}
    \label{tab:ratio}
\end{table}
}

\newcommand{\tabE}{
\begin{table}[t]
    \footnotesize
    \centering
    \renewcommand\arraystretch{1.0}
    \setlength{\tabcolsep}{4pt}
    \begin{tabular}{c|c|c|cc|cccc}
    $\alpha$ & $\#A$   & AP & AP$_{50}$ & AP$_{75}$ & AP$_{S}$ & AP$_{M}$ & AP$_{L}$ \\
    \headrule
    \multirow{3}*{$1.0$} & 1 & 40.4 & \textbf{59.6} & 43.9 & 23.5 & 43.5 & 53.7 \\
    &  2 & 40.4 & \textbf{59.6} & 44.1 & 23.7 & 43.4 & 53.6 \\
    &  3 & 39.9 & 59.4 & 43.3 & \textbf{23.9} & 43.2 & 51.9 \\ 
    \normrule
    \multirow{5}*{$1/3$}
    &  1 & 40.6 & 58.9 & 44.3 & 23.0 & 44.0 & 54.2 \\
    &  2 & 40.6 & 58.8 & 44.4 & 23.1 & 44.0 & 54.3 \\
    &  3 & \textbf{40.9} & 59.0 & 44.3 & 23.3 & 44.3 & 54.2 \\ 
    &  4 & 40.8 & 58.9 & \textbf{44.6} & 23.6 & \textbf{44.5} & \textbf{54.4} \\ 
    &  5 & 40.3 & 58.6 & 44.3 & 22.8 & 43.9 & 53.0 \\
    \end{tabular}
    \vspace{0.3em}
    \caption{\textbf{Tiling more anchors} when the regularizing term $\alpha=1.0$ and $\alpha=1/3$ in MuSu.\vspace{-1.0em}}
    \label{tab:anchor}
\end{table}
}

\newcommand{\tabSOTA}{
\begin{table*}[t!]
    \footnotesize
    \centering
    \renewcommand\arraystretch{1.0}
    \setlength{\tabcolsep}{10pt}
    \begin{tabular}{l|c|c|cc|cccc}
    method & backbone & AP & AP$_{50}$ & AP$_{75}$ & AP$_{S}$ & AP$_{M}$ & AP$_{L}$ \\
    \headrule
    FCOS~\cite{DBLP:conf/iccv/TianSCH19}
    & ResNet-101 & 41.5 & 60.7 & 45.0 & 24.4 & 44.8 & 51.6 \\
    FreeAnchor~\cite{DBLP:conf/nips/ZhangWLJY19}
    & ResNet-101 & 43.1 & 62.2 & 46.4 & 24.5 & 46.1 & 54.8 \\
    SAPD~\cite{DBLP:conf/eccv/ZhuCSS20}
    & ResNet-101 & 43.5 & 63.6 & 46.5 & 24.9 & 46.8 & 54.6 \\
    MAL~\cite{DBLP:conf/cvpr/KeZHYLH20}
    & ResNet-101 & 43.6 & 61.8 & 47.1 & 25.0 & 46.9 & 55.8 \\
    ATSS~\cite{DBLP:journals/corr/abs-1912-02424}
    & ResNet-101 & 43.6 & 62.1 & 47.4 & 26.1 & 47.0 & 53.6 \\
    AutoAssign~\cite{DBLP:journals/corr/abs-2007-03496}
    & ResNet-101 & 44.5 & 64.3 & 48.4 & 25.9 & 47.4 & 55.0 \\
     PAA~\cite{DBLP:journals/corr/abs-2007-08103}  
    &  ResNet-101 &  \textbf{44.8} &  \textbf{63.3} &  48.7 & \textbf{26.5} &  \textbf{48.8} &  56.3 \\
    MuSu (ours)    & ResNet-101 & \textbf{44.8} & 63.2 & \textbf{49.1} & 26.2 & 47.9 & \textbf{56.4} \\
    \midrule

    SPAD~\cite{DBLP:conf/eccv/ZhuCSS20}
    & ResNet-101-DCN & 46.0 & 65.9 & 49.6 & 26.3 & 49.2 & 59.6 \\
    ATSS~\cite{DBLP:journals/corr/abs-1912-02424}    
    & ResNet-101-DCN & 46.3 & 64.7 & 50.4 & 27.7 & 49.8 & 58.4 \\
     PAA~\cite{DBLP:journals/corr/abs-2007-08103}  
    &  ResNet-101-DCN &  \textbf{47.4} &  \textbf{65.7} &  51.6 &  \textbf{27.9} &   \textbf{51.3} &  \textbf{60.6} \\
    MuSu (ours)   & ResNet-101-DCN & \textbf{47.4} & 65.0 & \textbf{51.8} & 27.8 & 50.5 & 60.0 \\
    
    \end{tabular}
    \vspace{0.2em}
    \caption{\textbf{Comparison on COCO \texttt{test-dev} set} by different training sample selection methods with ResNet-101 and ResNet-101-DCN.\vspace{-0.8em}}
    \label{tab:sota}
\end{table*}
}
\begin{document}

%%%%%%%%% TITLE
\title{Mutual Supervision for Dense Object Detection}

\author{
    Ziteng Gao \quad \quad Limin Wang\textsuperscript{\Letter} \quad \quad Gangshan Wu \\
State Key Laboratory for Novel Software Technology, Nanjing University, China\\
}

\maketitle
\ificcvfinal\thispagestyle{empty}\fi

%%%%%%%%% ABSTRACT
\begin{abstract}
    The classification and regression head are both indispensable components to build up a dense object detector, which are usually supervised by the same training samples and thus expected to have consistency with each other for detecting objects accurately in the detection pipeline.
    In this paper, we break the convention of the same training samples for these two heads in dense detectors and explore a novel supervisory paradigm, termed as Mutual Supervision ({\em MuSu}), to {respectively} and {mutually} assign training samples for the classification and regression head to ensure this consistency.
    MuSu defines training samples for the regression head mainly based on classification predicting scores and in turn, defines samples for the classification head based on localization scores from the regression head.
    Experimental results show that the convergence of detectors trained by this mutual supervision is guaranteed and the effectiveness of the proposed method is verified on the challenging MS COCO benchmark.
    We also find that tiling more anchors at the same location benefits detectors and leads to further improvements under this training scheme.
    We hope this work can inspire further researches on the interaction of the classification and regression task in detection and the supervision paradigm for detectors, especially separately for these two heads. \footnote{Code will be available at \url{https://github.com/MCG-NJU/MuSu-Detection}.}
\end{abstract}
\blfootnote{ \Letter: Corresponding author (lmwang@nju.edu.cn).}
%%%%%%%%% BODY TEXT

\section{Introduction}
Object detection has been drawing interest from researchers for decades as one of the fundamental visual tasks in the computer vision community, especially with the rise of convolutional neural networks (CNNs).
The community has witnessed the fast evolution of both the methodology and the performance of detectors from region-based ones~\cite{DBLP:conf/iccv/Girshick15, DBLP:conf/nips/RenHGS15, DBLP:conf/iccv/HeGDG17, DBLP:conf/cvpr/CaiV18, DBLP:conf/cvpr/ChenPWXLSF0SOLL19, DBLP:journals/corr/abs-1901-01892}, to one-stage dense ones~\cite{DBLP:conf/iccv/LinGGHD17,DBLP:conf/cvpr/RedmonDGF16, DBLP:conf/eccv/LiuAESRFB16, DBLP:conf/iccv/TianSCH19, DBLP:journals/corr/abs-1904-07850, DBLP:conf/iccv/YangLHWL19} and then to end-to-end transformer-based detectors~\cite{DETR, DDETR}.
Among these methods, one-stage detectors, also known as dense detectors, are favored in terms of both the speed and accuracy, as well as the fast convergence due to their tiling anchors densely to cover objects of various scales and aspect ratios and directly predicting bounding boxes with labels with these anchors.

As detection task is about classifying and localizing simultaneously, object detectors are expected to produce bounding boxes with both correct classification labels and fine localization, and of course dense detectors are no exceptions. For a dense detector, these two tasks are usually done with specialized classification and regression heads. For the same input feature map from the backbone network, these two heads are expected to function differently: the classification head translates it into classification scores invariant with small shifts  while the regression head transforms it to shift-equivalent localizing offsets from anchors to bounding boxes, which incurs intrinsic inconsistency between these two tasks.

\figA

An accurate dense object detector is supposed to produce high-quality bounding boxes with correct labels, which requires that these two heads of different functionalities coordinate at the same spatial location of final outputs. In other words, converged detectors should ensure spatial consistency on where the maximum classification and localizing scores appear for an object. However, even for a converged detector, this goal is hard to achieve and the maximum classification score and the most accurate localizing box for an object frequently appear at different locations for a training image as the input image depicted in Figure~\ref{fig:inconsistency}. This inconsistency hurts the performance of final models in the current detection pipeline, especially in the process of the common post-processing non-maximum suppression (NMS), which only keeps the box with the maximum classification score among overlapping ones without the consideration of localizing accuracy. As a result, bounding boxes with finer localization but lower classification scores are suppressed and such detectors lead to inferior performance.

To tackle this problem, previous work focuses on input features and network structures of these heads and disentangles the classification and regression heads from feature or structural perspectives. Different from those, we delve into this problem from the view of the supervision for these two heads, specifically, the definition of the training samples respectively for both them, and alleviate this inconsistency by proposing mutual supervision (MuSu) for dense detectors.

MuSu separates the definition of training samples for classification and regression head and then makes them dependent on each other mutually. As illustrated in Figure~\ref{fig:method}, training samples are not shared between two heads. Training targets for classification are adaptively determined by IoU (Intersection over Union) scores between predicted boxes and ground-truth boxes from the regression head. Alike, training samples for the regression head are defined by classifying scores from the classification head. Next, MuSu translates scores of these training samples for these two heads to soft targets by associating weights to losses of each spatial location. By this means, MuSu aims to force the consistency between these two heads by the mutual assignment in the training phase.
Under this mutual supervision scheme, MuSu also enjoys the advantage of the training samples adaptively emerging from the network itself, which are refrained from being hand-crafted by expert knowledge.
Moreover, MuSu is exempt from any hand-crafted geometric prior and also get rids of subtle treatments to different pyramid levels. In this sense, MuSu makes a big step further to fully adaptive sample assignment and unleashes the power of a detector more comprehensively.

We carry out extensive ablation experiments on MS COCO dataset~\cite{DBLP:conf/eccv/LinMBHPRDZ14} to validate the effectiveness of our proposed MuSu method. In particular, MuSu boosts the FCOS detector with ResNet-50 backbone to a 40.6 AP in the COCO validation set under the common 90k training scheme without the sacrifice of the inference speed. Moreover, we investigate that tiling more anchors at the same location will benefit the detector under this mutual supervision scheme, pushing to 40.9 AP over the competitive one-anchor counterpart. We argue that our method of mutual supervision for the classification and regression head exploits multiple anchor settings more fully and thus boosts the performance higher, in contrast to \cite{DBLP:journals/corr/abs-1912-02424}. We also utilize MuSu to train models with large backbones to compare state-of-the-art models and our models achieve promising results on COCO \texttt{test-dev} set. 

\section{Related Work}
\subsection{In the Context of Classification and Regression Heads}
The classification and regression head as sibling heads serve as essential components for general object detectors, where input features from the backbone network are transformed into classification scores and predicted boxes, respectively. Regional CNN (R-CNN) detectors~\cite{DBLP:conf/iccv/Girshick15, DBLP:conf/nips/RenHGS15, DBLP:conf/cvpr/CaiV18} commonly deployed the shared head (\textit{2fc}) in the regional network to classify and do the finer localization based on the region of interest (RoI) which is pooled out of the feature map.
The work \cite{DBLP:conf/eccv/ChengWSFXH18, DBLP:conf/cvpr/0008CYLWL020} proposed different heads for R-CNN detectors and disentangle them by the individual network to achieve consistency between the classification and regression output.
TSD~\cite{DBLP:conf/cvpr/SongLW20} argued that classification and regression heads need different spatial features and the shared RoI pooling operator is a cause to the misalignment.
For dense object detectors, things are different and not so straightforward to deal with for there are not RoI operators and the feature into different heads is hard to disentangle.
As a common practice in~\cite{DBLP:conf/iccv/LinGGHD17, DBLP:conf/iccv/TianSCH19}, the classification and localization heads are respectively comprised of several convolutional layers with the hope for different functionalities where the input feature is the same.

Different from previous work on the feature or the structure, our proposed method tackles the problem of inconsistency from the perspective of designing training samples respectively for each head.
Previous methods on the supervision~\cite{DBLP:conf/eccv/JiangLMXJ18, DBLP:conf/iccv/TanNQLL19, PISA, varifocal} involved solely the unidirectional supervision either from the regression to the classification or vice versa.
In contrast, our proposed MuSu supervises each head by training samples defined by the counterpart head output and ensures the consistency in a bidirectional manner.
The work most related to this paper~\cite{MutualGuidance} shares the same address with our method, which defined the customized IoU criteria with the consideration of the counterpart head output for sample definition. However, details in~\cite{MutualGuidance} are highly hand-crafted with the customized IoUs and the improvement is not validated on recent detectors while our MuSu brings improvements over strong baselines with the simple and adaptive supervision scheme design.

\figB

\subsection{In the Context of Training Sample Selection}
The most popular strategy to select training samples is to use IoU as a criterion between an anchor and a ground-truth box, dating back to \cite{DBLP:conf/cvpr/GirshickDDM14, DBLP:conf/nips/RenHGS15}.
Recently, various training sample selection strategies are proposed based on either the geometric relation, the classification score, the IoU or jointly them, to determine which object an candidate anchor belongs to in the training phase, and exploit the potential of a detector more further. FreeAnchor~\cite{DBLP:conf/nips/ZhangWLJY19} was the first to adaptive training samples based on the customized likelihood by classification scores and IoUs. The literature~\cite{DBLP:conf/cvpr/KeZHYLH20, DBLP:conf/cvpr/LiWZXSD20, DBLP:journals/corr/abs-2007-03496, OneNet} proposed to explicitly select training samples by the joint criteria of the classification and the regression. ATSS~\cite{DBLP:journals/corr/abs-1912-02424} utilized the statistics of IoUs with regard to anchors with an object to determine positive samples. PAA~\cite{DBLP:journals/corr/abs-2007-08103} introduced a probabilistic process to the training sample selection and determine samples by the expectation-maximization algorithm.
All these work cast improvements over the performance and has indicated the significance of designing better training samples.

Our method follows this research line but differs from these methods above. We step further in this line of adaptive training samples by assigning different samples to different heads and our proposed method automatically mines classification samples from the IoU and regression samples from the classification score. 
Fortunately, with this mutual supervision, our method MuSu also gets rid of the geometric prior and subtle treatment for each pyramid level in these adaptive approaches and in that sense, our proposed MuSu method is the neatest way to adaptively assign training samples by far while achieving promising results.

\section{Proposed Method}

To make accurate detections, a dense detector is expected to have alignment between the classification and regression heads since that the post-processing NMS only keeps detections with the maximum classification confidence when there are multiple overlapping ones. In detectors like  RetinaNet~\cite{DBLP:conf/iccv/LinGGHD17}, the classification head is trained by the supervision signal where the overlapping of predicting and ground-truth box is higher over a certain value, without the further consideration of how well the ground-truth is localized.
Indeed, the current pipeline expects that the classification confidence represents not just how well the detector classify but also how well the detector localizes, as argued by \cite{DBLP:journals/corr/abs-2006-04388, DBLP:journals/corr/abs-2007-08103, DBLP:conf/eccv/JiangLMXJ18, DBLP:conf/iccv/TianSCH19}. Therefore, the spatial distribution of the supervision for the classification head should rely on the localizing performance of the regression head, that is, where the IoU score is larger, where the classification supervision is stronger.
In turn, the supervision for the regression should be also imposed more on the positions with higher classification scores, forcing well-classified ones to regress accurately as well.
Figure~\ref{fig:method} depicts this dependency between them and mutual supervision. 

We introduce the {\em Mutual Supervision} (MuSu) algorithm for dense object detectors as a simple instantiation of this mutual philosophy. Specifically, MuSu ensures the consistency between the classification and regression in the training procedure by assigning training samples mutually and reciprocally from and for these two heads. MuSu treats the training sample in the soft target form by weighting losses of anchors in a ranking mechanism.
MuSu can be described as three steps:
\textbf{i)} construct adaptive candidate bags jointly by the classification and regression head to select the candidate anchors most belonging to an object;
\textbf{ii)} compute candidate rankings from the perspectives of the classification and regression respectively inside the candidate bags;
\textbf{iii)} translate these rankings to weights to sum out losses of each position and supervise the classification and regression head. The MuSu algorithm is depicted in Procedure~\ref{algo}.

\subsection{Adaptive Candidate Bag Construction}\label{sec:bag}
We first construct adaptive candidate bag for every object as a preliminary step to filter out plenty of unsuitable anchors to better perform the following mutual supervision. The proposed candidate bag adaptively keeps out the false candidates to an object jointly by the classification and regression head and prevent anchors which obviously belong to the background or other objects from feeding into the next procedure. Otherwise, the mutual scheme will confuse detectors since classification scores are instance-agnostic and tend to be noisy at the initial stage.

More formally, considering for an object $j$, given an anchor $i$ as well as its classification score $p_i$ and IoU criterion $\text{IoU}_i$ w.r.t. object $j$, we define a joint likelihood of weighting how much an anchor $i$ is a candidate to object $j$, that is
\vspace{-0.1em}
\begin{equation}\label{equ:joint}
    P_i = p_i q_i, \vspace{-0.0em}
\end{equation}
where $q_i=\text{IoU}_i^\theta$ and $\theta$ is the weighting coefficient that rescales IoU in the exponential way to approximate the range of classification scores. $\theta$ is set to $4$ our experiments. We calculate the threshold of candidate bags by
\vspace{-0.1em}
\begin{equation}\label{equ:threshold}
    t = b\cdot \max_i P_i, \vspace{-0.0em}
\end{equation}
where $b$ is the thresholding parameter less than $1$.
Any anchor in the ground-truth box with the joint likelihood higher than $t$ becomes a candidate in the candidate bag $\mathcal{C}_j$ for an object $j$. As this procedure only chooses candidates loosely and filters out obviously unsuitable anchors, the parameter $b$ is preferred to a low value, \eg, $0.1$.
For multiple objects situation, we only keep the object $j$ with the highest IoU in an anchor $i$ involved in this computation and leave other ground-truth boxes out of consideration. This also makes the candidate bag mutually exclusive with regard to the object. By this means, we assign the ground-truth box $j$ to each candidate $i\in\mathcal{C}_j$ without conflicts.

The candidate bag is also adaptive in its size. When the misalignment between the classification and regression head occurs, the threshold of a candidate bag becomes lower as the maximum of the joint likelihood for an object is also lower. More candidates are selected into the bag under this situation and this mechanism enables us to mines hard objects concerning the inconsistency between the classification and regression by enlarging their sample number and focus on them during training.

\algomusu

\subsection{Mutual Supervision}
As we complete constructing candidate bags to filter out background anchors, we are ready to apply our proposed mutual supervision for classification and regression heads. For each head, MuSu assigns each candidate a ranking in descending order by evaluating the accuracy between the counterpart head prediction and the ground-truth object. Then MuSu translates the ranking to the weight for each candidate. We reuse the classification score $p_i$ and the scaled IoU $q_i$ in Equation~\ref{equ:joint} as the evaluation criteria for the classification and regression head. A natural choice is to use $p_i$ for computing rankings for candidates of the regression head, $R^{\mathit{reg}}_i$, and use $q_i$ for rankings of the classification head, $R^{\mathit{cls}}_i$. However, we found that this straightforward way of mutual supervision performed inferior in our experiments. In contrast, MuSu utilizes the regularized criteria values to compute rankings for candidates in the symmetric form,
\vspace{-0.1em}
\begin{align}\label{equ:value}
\begin{cases}
    v^{\mathit{cls}}_{i}=q_i \cdot p^{\alpha}_i,\vspace{0.5em}\\
    v^{\mathit{reg}}_{i}=p_i \cdot q^{\alpha}_i,
\end{cases}
\end{align}
 where $\alpha$ acts like a hyper-parameter, varying from $0$ to $1$, which regularizes the mutual scheme
by also considering the output of the head itself.
Our mutual supervision scheme can be a generalized training sample framework, where $\alpha=1$ gives recently studied training sample strategies based on joint likelihood by these two heads~\cite{DBLP:journals/corr/abs-2007-03496, DBLP:conf/cvpr/KeZHYLH20, DBLP:journals/corr/abs-2007-08103} and $\alpha=0$ gives the straightforward mutual supervision without the regularization of the supervised head itself.
 
\subsection{Loss Weighting Paradigm}
As we obtain regularized criteria $v^{\mathit{cls}}_{i}$ and $v^{\mathit{reg}}_{i}$, MuSu sorts these values in a descending order within each candidate bag separately for both the classification and regression head to acquire the ranking $R^{\mathit{cls}}_i$ and $R^{\mathit{reg}}_i$ (starting from $0$, increasing by step size $1$, \ie, $0, 1, 2, \cdots$). MuSu supervises these two heads in the soft target form by weighting losses for each candidate and summing these weighted losses into a total loss. The weights $w^{\mathit{cls}}_i$ and $w^{\mathit{reg}}_i$ for candidates are decided by the ranking of each candidate separately for two heads and MuSu adopts a negative exponential way to translate rankings to weights:
\vspace{-0.1em}
\begin{equation}
\label{equ:translate}
\begin{cases}
    w^{\mathit{cls}}_{i} = \exp(-R^{\mathit{cls}}_i/\tau^{\mathit cls}),\vspace{0.5em}\\
    w^{\mathit{reg}}_{i} = \exp(-R^{\mathit{reg}}_i/\tau^{\mathit reg}),
\end{cases}
\end{equation}
\vspace{-0.1em}
where $\tau^{\mathit{cls}}$ and $\tau^{\mathit{reg}}$ are temperature coefficients for the classification and regression head, indicating how many weights are assigned for samples to an object. As the ranking $R^{(\cdot)}_{i}$ increases ($v^{(\cdot)}_i$ becomes smaller), the weights are exponentially decreasing at a speed related to the temperature $\tau^{(\cdot)}$. Thanks to the mutual supervision scheme, we can control the number of positive training samples respectively for each head and we found that the performance would be better if we assign less weights to the regression head.

The total loss in an image for each head can be formulated in general as:
\begin{equation}\label{equ:total_loss}
    \mathcal{L}=\frac{1}{N}\sum_{i}w_i\ell_i,
\end{equation}
where the normalized term $N=\sum_{i} w_i$ and $\ell_i$ is the loss function with regard to the prediction and ground-truth assigned for each anchor $i$.
$\ell$ can be arbitrary loss functions for each head, \eg, the focal loss~\cite{DBLP:conf/iccv/LinGGHD17} for classification and GIoU~\cite{DBLP:conf/cvpr/RezatofighiTGS019} loss for regression. Details with not-assigned classes for the focal loss are discussed in Section~\ref{details}.

It is notable that MuSu is not a specific loss function for either classification or regression and \textit{de facto} acts as a hyper-loss formulation built upon these base losses. Actually, the focus of MuSu is to discuss the sample assignment for each head in two aspects: first, the ground-truth assignment for position $i$, stands for which object is the target of position $i$ to supervise; second, weights for assigned training samples, $w_{i}$, indicating how much the position $i$ should be supervised.
As a plus, we separate the assignment strategies from underlying loss function choices and put attention on the relative rankings of anchors inside candidate bags, guaranteeing that the absolute amplitude of losses has no effect on assignments.
We summarize our proposed mutual supervision approach MuSu as several key points: {\em first},
MuSu exploits the spatial distribution of scores originating from the counterpart head to adaptively determine training samples for the classification and regression heads in a respective manner.
This paradigm circumvents either any hand-crafted training sample assignments or geometric clues\footnote{Except for the inner-box restriction, which is necessary as the FCOS-like architecture uses non-negative distance to predict offsets to bounding box borders.}. Thus, MuSu emerges as a simple and general training sample selection approach;
{\em second},
MuSu enables detectors to align classification scores to IoUs scores, making detectors friendly to the NMS procedure and the final detection evaluation;
{\em third},
MuSu disentangles the training sample assignment and the choices of base loss functions in the mutual supervision since it utilizes the relative ranking to determine loss weights associated with these anchors, which is extensible to any loss function improvement in the future;
{\em finally}, MuSu alleviates the regressing difficulty of the regression head by assigning fewer positive samples for it and focusing on positions with higher classification scores.
Experiments shows that MuSu-trained detectors lead to the consistent better performance.

\section{Experiments}
To validate the effectiveness of our proposed mutual supervision scheme for classification and regression head, we conduct experiments on MS COCO detection dataset~\cite{DBLP:conf/eccv/LinMBHPRDZ14} in this section. Following the common practice of previous work, we use \texttt{trainval35k} subset consisting up of 115K images to train our models and use \texttt{minival} subset of 5K images as the validation set. We report our ablation study results on \texttt{minival} subset. We also submit our final model results on the \texttt{test-dev} subset, whose labels are not publicly visible, to the MS COCO evaluation server to compare with state-of-the-art models. We implement our MuSu method in mmdetection codebase~\cite{mmdetection2018}.
\subsection{Implementation Details}\label{details}

\textbf{Network structure.}
Theoretically, our mutual supervision method is universal for dense object detectors. In this paper, we adopt the recently-proposed dense detector FCOS~\cite{DBLP:conf/iccv/TianSCH19} as our network architecture. The FCOS architecture serves as a strong baseline for dense detectors by utilizing Group Normalization~\cite{DBLP:journals/ijcv/WuH20} to both the classification and regression detection head, adding trainable scalars for each pyramid level on FPN~\cite{DBLP:conf/cvpr/LinDGHHB17} and using the centerness layer from the last feature map of the regression head to filter out a number of inaccurate detections. As our proposed method selects training samples adaptively and does not depend on the fixed centerness estimation, following previous work~\cite{DBLP:journals/corr/abs-2007-03496}, we redirect the output of the centerness layer in the FCOS architecture to the output of the classification head as so-called implicit objectness and merge them by multiplication to get final classification scores. 

\textbf{Initializations.}
All backbones of detectors throughout our experiments are initialized from the pre-trained models on the ImageNet dataset~\cite{DBLP:conf/cvpr/DengDSLL009}. For stabilization during early training, we initialize weights of the last convolutional layer in the regression head to zeros. We also set a constant stride factor on each feature pyramid level of FPN~\cite{DBLP:conf/cvpr/LinDGHHB17} to scale regression boxes, starting from stride $s=8$ at the finest pyramid level $P_3$ to $s=128$ at the level $P_7$. These settings make boxes predicting from the regression branch for each position initialized to the same size $2s\times 2s$ for a FPN level, serving as a geometric prior in early iterations for more stable mutual supervision.

\textbf{Mutual supervision instantiation.}
We set the temperature $\tau^{\textit{cls}}$, which controls the number of positive samples assigned to an object, to the squared root of the candidate bag size, and then set the temperature for the regression branch to the half of the temperature for the classification ($\tau^{\textit{cls}}:\tau^{\textit{reg}}=2:1$) as our default. That is,

\begin{equation}\label{equ:temp}
\begin{cases}
\tau^{\textit{cls}}=\sqrt{|\mathcal{C}_j|},\vspace{0.2em}\\
\tau^{\textit{reg}}=0.5 \tau^{\textit{cls}}=0.5\sqrt{|\mathcal{C}_j|}.
\end{cases}
\end{equation}

The temperature $\tau^{\textit{cls}}$ and $\tau^{\textit{reg}}$ are specific to a candidate bag of a ground-truth object $j$. The squared root operator makes the temperature vary moderately across different objects when the candidate bag size varys a lot and leads to more stable training. We set the thresholding coefficient $b$ in Equation~\ref{equ:threshold} to $0.1$ as our default.

We adopt the focal loss~\cite{DBLP:conf/iccv/LinGGHD17} as our base loss for the classification and the GIoU loss~\cite{DBLP:conf/cvpr/RezatofighiTGS019} for the regression. The focal loss tackles the classification task in detection as the multi-class binary classification problem. For an anchor, there exists the negative classification of non-target classes along with the positive classification of the target class. In addition, the negative classification should be also applied to the assigned class with soft targets. Thus, we treat it carefully and separate the focal loss into three parts: the positive term for the assigned class label, the negative penalty term for the assigned class label, and the background term for all other not-assigned class labels. We extend the loss form in Equation~\ref{equ:total_loss}:
\vspace{-0.1em}
\begin{align}
    \mathcal{L}_{\mathit{cls}}=\frac{1}{N}\sum_{i}[
    {w^{\mathit{cls}}_i\cdot \ell^{+}_{i}}
    +
    (1-w_i^{\mathit{cls}})^\beta\cdot\ell^{-}_{i}
    +
    \ell^{\mathit{bg}}_i
    ],
    \vspace{-0.1em}
\end{align}
where the $\ell^{+}_i$ is the focal loss to the positive classification of the assigned label while $\ell_i^{-}$ is the focal loss for the negative classification of the assigned label as the penalty for the position with insufficient $w^{\mathit{cls}}_i$. These two loss terms function as the soft target in a unified manner. Background loss term $\ell^{bg}_i$ is the sum of the focal loss for negative classification of all other classes which are not assigned to anchor $i$. The focusing and balance parameter for the focal loss follows the default settings in \cite{DBLP:conf/iccv/TianSCH19} and the penalty decay term $\beta$ is set to $4$ following \cite{DBLP:conf/eccv/LawD18, DBLP:journals/corr/abs-1904-07850}. The total loss for detection is simply the sum of the classification and regression loss,
\begin{align}
    \mathcal{L}_{\mathit{det}}=\mathcal{L}_{\mathit{cls}}+\mathcal{L}_{\mathit{reg}}.
\end{align}

\textbf{Optimization and inference.}
We use SGD with the learning rate 0.01, momentum factor 0.9, and weight decay 0.0001 to optimize our models throughout experiments. A total batch of 16 images, 2 images per GPU, are used in the training. The statistics and affine parameters of batch normalization layers in the backbone are frozen as in \cite{DBLP:conf/iccv/TianSCH19}. For ablation studies, we train models with the ResNet-50 backbone~\cite{DBLP:conf/cvpr/HeZRS16} in 90K iterations with the learning rate warmup during the first 500 iterations. The learning rate is divided by a factor of $10$ at the 60K and 80K iteration, respectively. All images in the 90K training scheme are resized to their shorter size being $800$ and their longer size not greater than $1333$ and are randomly flipped horizontally as the only data augmentation. At the inference stage, we resize the input image to the same size in the training procedure without random flipping. The threshold of the classification score is set to $0.05$ and the NMS threshold is set to $0.6$ in the detection pipeline, also following recent common practice. The optimization and inference details are kept the same throughout our experiments unless otherwise stated.

\subsection{Training with Mutual Supervision}
\tabA

\textbf{Study on the mutual supervision}.
We start our experiments from the vanilla FCOS detector as our baseline. The vanilla FCOS is supervised by dense signals and serves as a competitive baseline for dense detectors, which got 36.5 AP in Table~\ref{tab:effective}. The FCOS++ model in Table~\ref{tab:effective} denotes the improved architecture and more importantly, the refined highly hand-crafted training sample, which only assigns positive samples within the center area of objects.
In contrast, our MuSu emerges as an adaptive training sample assignment approach for dense object detectors and the key component in MuSu is the criteria value in Equation~\ref{equ:value} as it decides which training sample selection strategy for each head our method adopts. 
In Table~\ref{tab:effective}, we carry out experiments of various settings for criteria values in Equation~\ref{equ:value}. 
The settings whose weights for two heads are totally decided by the single head output (the classification $p_i$ or the regression $q_i$) and assign the same criteria value for both two heads, which in other words enables the sole unidirectional supervision without the mutual scheme, leads to inferior results of 38.3 and 31.8 AP respectively.

\tabB
\tabC
\tabD

The naive mutual supervision without the regularized term (setting $\alpha=0.0$) achieves 38.5 AP, comparable to the highly hand-crafted FCOS++ model.
When we add the regularized factor, even from $\alpha=1/6$, the performance of models significantly improves to 40.4 AP.
With the regularized factor $\alpha=1/3$, the performance of models trained with the mutual supervision leads to a best 40.6 AP, 2.0 AP higher than the FCOS++ model. We argue that the regularizing term is necessary for the assignment because it is also aware of \emph{how well each head itself learned} in the training procedure and makes the best of the use of each head prediction to avoid the assignment fluctuation. It is notable that the MuSu with $\alpha=1.0$, which assigns the same criteria value for training sample selection for two heads, is also a case of mutual supervision where the supervision of a head is also aware of the prediction of the counterpart head. In this sense, we include training samples based on the joint likelihood explored by the recent approach~\cite{DBLP:journals/corr/abs-2007-03496, DBLP:conf/cvpr/KeZHYLH20, DBLP:journals/corr/abs-2007-08103} in our proposed MuSu approach. However, this same criteria strategy ($\alpha=1$) suffers from stagnant or even degenerate performance when tiling more anchors while the MuSu with $\alpha=1/3$ benefits from more anchors as discussed below.

\textbf{Study on adaptive candidate bags and temperature}.
As we discuss in Section~\ref{sec:bag}, the candidate bag is designed for filtering out the plenty of background anchors adaptively by the joint likelihood of the classification and regression. The candidate bag only serves as a preliminary procedure to keep obviously unsuitable anchors from the next mutual assignment procedure, so the threshold coefficient $b$ is preferred to a relatively low value. In Table~\ref{tab:anchor_bag}, we vary the coefficient to see its impact. The coefficient $b=0.10$ gives the best result.

A candidate bag is also adaptive with regard to its size. On account of that, MuSu can put more focuses on objects with strong inconsistency between classification and regression by assigning more positive samples to them in the relation depicted in Equation~\ref{equ:translate} and \ref{equ:temp}. We validate the effectiveness of adaptive candidate bag on final detectors by disabling the adaptive temperature w.r.t the bag size and setting $\tau^{\mathit{cls}}$ and $\tau^{\mathit{reg}}$ to a fixed number. Borrowing the average temperature $\tau^{\mathit{cls}}$ and $\tau^{\mathit{reg}}$ when using adaptive candidate bags, the temperature for the classification $\tau^{\mathit{cls}}$ is set to the fixed constant $5.0$ across objects and keep $\tau^{\mathit{cls}}:\tau^{\mathit{reg}}=2:1$. For more ablation, we add the situation $\tau^{\mathit{cls}}=10.0$ in Table~\ref{tab:temperature}. We find that the adaptive temperature as the function of the bag size benefits our method by adaptively mining hard objects with regard to the inconsistency.
We also present results of applying different ratios of temperatures for assigning samples to each head ($\tau^{\mathit{cls}}$:$\tau^{\mathit{reg}}$) under the setting of adaptive temperature to anchor bag sizes in Table~\ref{tab:ratio}, which indicates that moderately reducing samples for regression is favourable for the fine localization and overall performance.

\tabE

\tabSOTA
\textbf{Soft versus hard targets}.
As our method defines training samples for both heads as soft targets by weighting losses of them, one natural question is whether we can use hard targets instead of soft targets to achieve a similar performance. We trained the model with hard targets by modifying Equation~\ref{equ:translate} to $w^{(\cdot)}_i=\mathbb{I}[R^{(\cdot)}_i<\tau^{(\cdot)}]$, where $\mathbb{I}[\cdot]$ is the indicating function and this resulted in a 40.0 AP model, which is 0.6 AP behind the soft target scheme. This comparison supports the conclusion of the literature \cite{DBLP:journals/corr/abs-2006-04388, DBLP:conf/eccv/ZhuCSS20, varifocal} and soft targets in our method share the same idea in the flexible classification to align regression scores.

\textbf{Tiling more anchors}.
Placing multiple anchors at each spatial position of output detection maps is a common way to cover image boxes of different scales and aspect ratios as many as possible in dense object detectors. This strategy is popular among both one-stage detectors~\cite{DBLP:conf/iccv/LinGGHD17} or proposal networks of two-stage detectors~\cite{DBLP:conf/nips/RenHGS15} for achieving better performance. However, recent work~\cite{DBLP:conf/iccv/TianSCH19, DBLP:journals/corr/abs-1912-02424} challenges this necessity of tiling more anchors by changing sample assignment strategies and shows that there are no performance gains by placing more anchors under their settings.

To discuss multiple anchor situations, we set the initial scale and aspect ratio of anchors by initializing the bias parameter of the last convolutional layer to produce bounding boxes. The scale factor of an anchor and the aspect ratio are drawn uniformly at random from the interval $[1, 2]$ and $[\frac{1}{2}, 2]$, respectively.

Surprisingly, we find that tiling more anchors has a boost on the detection performance over competitive results under our mutual supervision scheme, even without well-crafted settings of scales and aspect ratios. As shown in Table~\ref{tab:anchor}, these results show that MuSu enables the detector to fully exploit the setting of more anchors. The performance of a detector can increase to about 40.9 AP when adding anchor per location to 3 or 4. In contrast, the counterpart results, which assigns the same criteria values for two heads with $\alpha=1.0$, will not be better when adding more anchors and even suffer from that.
The final MuSu model is 2.3 AP higher than FCOS++ model, 4.1 AP higher than the vanilla FCOS model, and 0.5 AP higher than our competitive baseline with $\alpha=1.0$ and $\#A=1$. This empirical evidence validates the effectiveness of our MuSu method beyond the single anchor situation.
\subsection{Comparison to the State of the Art}
To compare with other state-of-the-art methods of training sample selection for detection, we use deeper backbones and deformable one~\cite{DBLP:journals/corr/abs-1811-11168} to train with our MuSu. To align with previous work and compare fairly, we extend the training schedule to the 180K iterations and reduce the learning rate at the 120K and 160K iteration by a factor of $0.1$. For an input image, we resize the shorter side to a scale randomly chosen value of $[640, 800]$. We train our MuSu detectors with 3 anchors per location ($\#A=3$). For the DCN variant, we also apply deformable convolutional layer to the last layer on each head following~\cite{DBLP:journals/corr/abs-1912-02424, DBLP:journals/corr/abs-2007-08103}. As shown in Table~\ref{tab:sota}, both the ResNet-101 detector and the DCN variant trained by the MuSu surpass previous competitive models in the overall AP while achieving the new state-of-the-art AP$_{75}$ without bells and whistles at the inference stage.
Further, MuSu-trained models are on par with PAA models that are with the score voting as the improvement at inference stage.

It is worth noting that our MuSu offers as a simple instantiation of our proposed mutual supervision and this scheme, in general, is also compatible to specific training sample selection methods, such as the PAA algorithm~\cite{DBLP:journals/corr/abs-2007-08103} for each head to expect better results.

\vspace{-0.3em}
\section{Conclusion}
In this paper, we have presented the mutual supervision (MuSu) scheme for training accurate dense object detectors in which we break the convention of the same training samples for the classification and regression heads and then these two heads are supervised based on the output of each other in the soft target way. MuSu makes a big step further to fully adaptive training sample selection by means of assigning different samples to these two heads in a mutual manner without the subtle geometric designing. Moreover, we discuss multiple anchor settings under our proposed mutual supervision and find that is beneficial to our method. Experimental results on the challenging MS COCO benchmark validate the effectiveness of our proposed MuSu training scheme on detectors. 

\vspace{-0.3em}
\paragraph{\bf Acknowledgements.} This work is supported by National Natural Science Foundation of China (No. 62076119, No. 61921006), Program for Innovative Talents and Entrepreneur in Jiangsu Province, and Collaborative Innovation Center of Novel Software Technology and Industrialization.

\clearpage
\newcommand{\AppendTabA}{
\begin{table}[t]
    \small
    \centering
    \renewcommand\arraystretch{1.0}
    \setlength{\tabcolsep}{6pt}
    \begin{tabular}{l|cccc}
    models & \#Params & FLOPs & FPS \\ 
    \headrule
    FCOS
      & 32.07M & 205.3G & 14.6\\
    \normrule
    MuSu ($\#A=1$)
      & 32.07M & 205.3G & 14.6\\
  MuSu ($\#A=2$)
      & 32.27M & 209.5G & 13.2\\
    MuSu ($\#A=3$)
      & 32.46M & 213.7G & 12.5\\
    MuSu ($\#A=4$)
      & 32.66M & 217.8G & 11.7\\
    MuSu ($\#A=5$)
      & 32.85M & 222.0G & 11.0\\
    \end{tabular}
    \vspace{0.2mm}
    \caption{Comparisons of parameters and FLOPs with the backbone ResNet-50. The FLOPs of models are calculated with the input shape $(1333, 800)$. The FPS is measured by a single Titan Xp.}
    \label{tab:AppendFlops}
  %  \vspace{-2.5em}
\end{table}
}

\newcommand{\AppendFigA}{
    \begin{figure}[H]
        \centering
        \includegraphics[width=.7\textwidth]{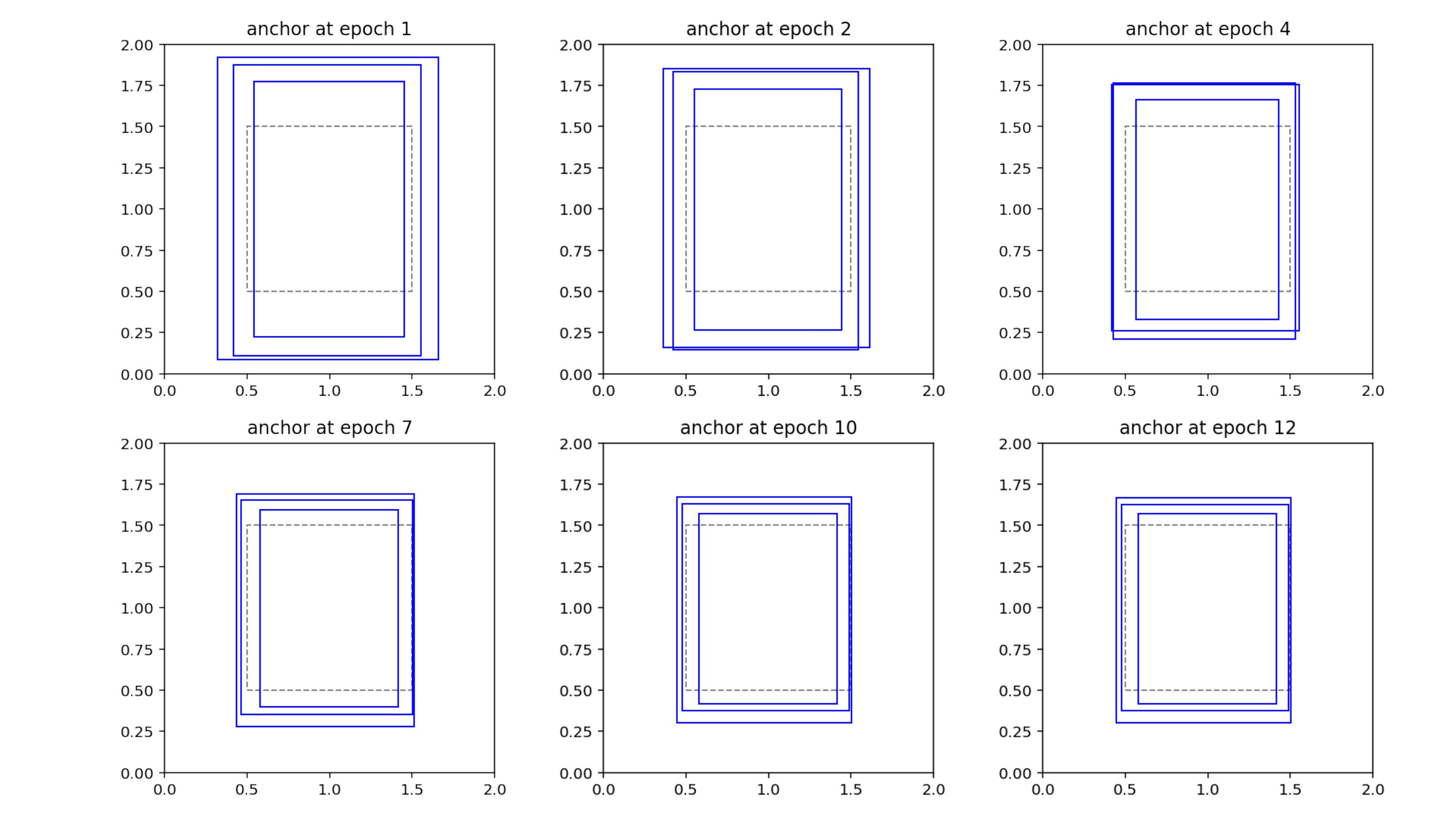}
        \caption{Multiple anchors visualized by the bias parameter of the last convolution to produce bounding boxes during the training procedure by the MuSu. Dashed boxes stand for basic spatial units on the feature map. One can see, when converged, anchors are specific to different scales and aspect ratios even though during training (\eg, at the epoch 4) several anchors are similar. Another key ingredient of multiple anchor settings, the weight parameter of the last convolution layer to bounding boxes, also can lead to specific preference to different scales and aspect ratios. }\label{fig:AppendAnchors}
    \end{figure}
}

\newcommand{\AppendFigC}{
    \begin{figure}[b]
        \centering
         \includegraphics[width=0.45\textwidth, trim=0 0 0 0,clip]{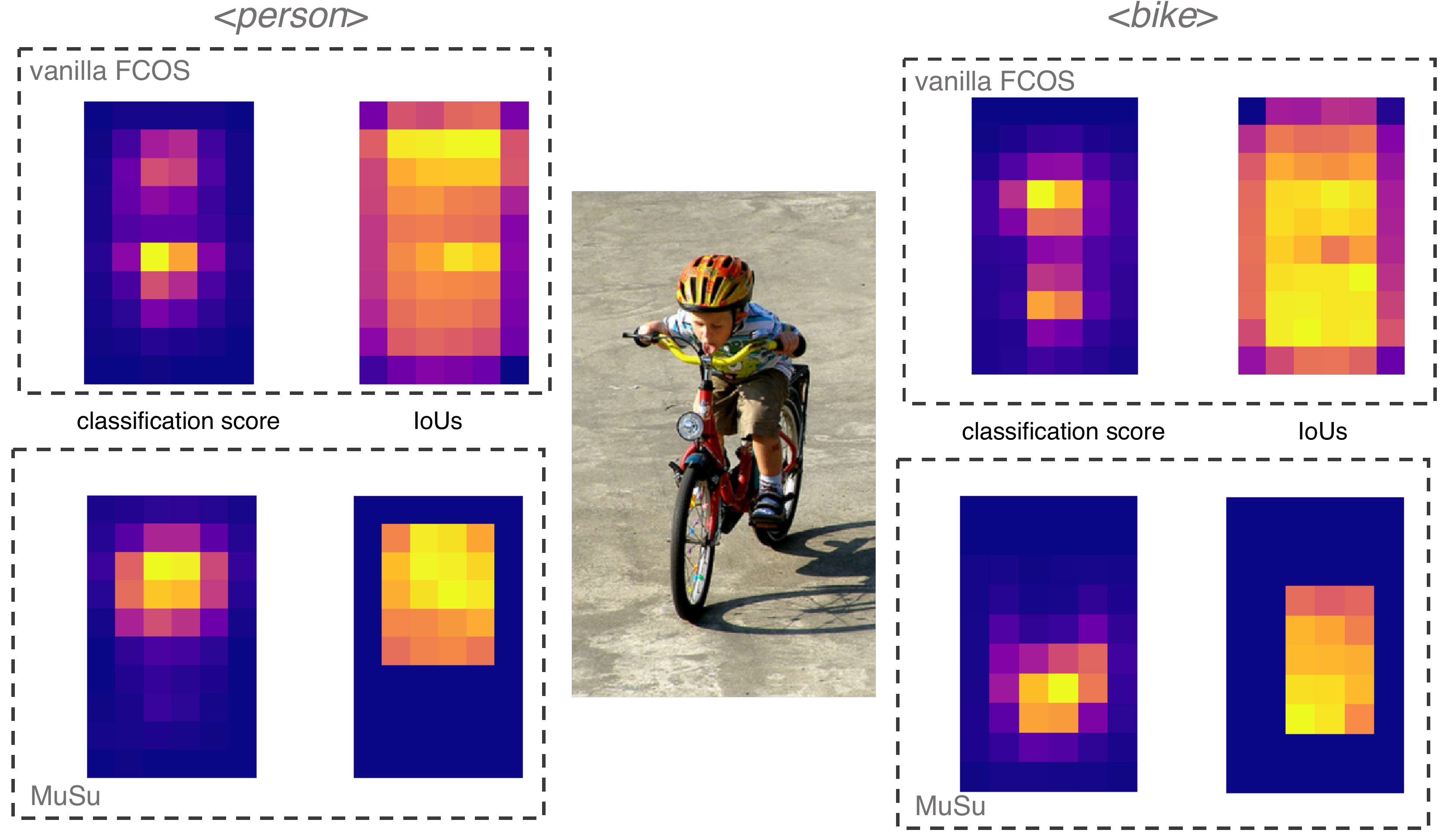}
    \caption{More visualizations of the consistency between heads. Note that the vanilla FCOS also faces the entanglement of classification scores of different classes, which our MuSu is exempt from. }\label{fig:AppendMore}
    % \vspace{-1em} 
    \end{figure}
}

\newcommand{\AppendFigE}{
\begin{figure*}[t]
\centering
\includegraphics[width=\textwidth]{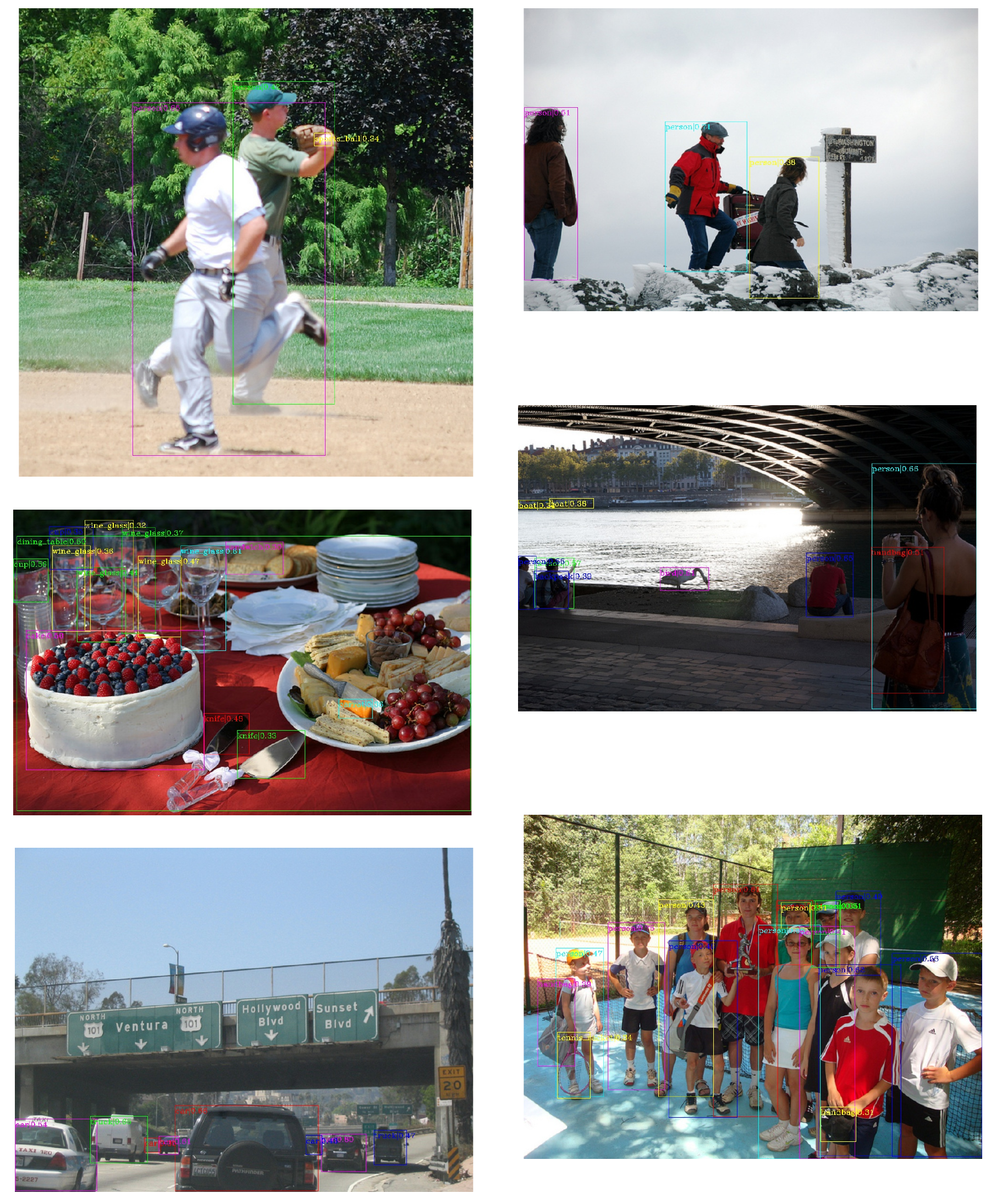}
\caption{Visualization of detection results of the MuSu model with $\#A=3$ and R-50 backbone on COCO \texttt{minival} split. Better zoom in.  \vspace{-0.5em}}
\label{fig:det}
\end{figure*}
}

% \AppendFigA
\twocolumn[{%
\renewcommand\twocolumn[1][]{#1}%
% \maketitle
\begin{center}
% hhh
\begin{minipage}[t]{\linewidth}
\AppendFigA
\end{minipage}
\end{center}%
}]
% \AppendFigA

\section*{Appendix I. Details of Learned Anchors}
In the paper, we showed that multiple anchor settings benefit the MuSu supervision scheme and MuSu actually enables the network to group anchors into different scales and aspect ratios. Here, we show details of learned anchors when $\#A=3$ trained by MuSu in Figure~\ref{fig:AppendAnchors}.

\section*{Appendix II. Computational Complexity}
It is worth noting that our MuSu models with 1 anchor per location ($\#A=1$) on the feature map share the \textbf{exact same} number of parameters and FLOPs with the FCOS models since we do not add any new modules for MuSu models. For the multiple anchor variants, the parameters and FLOPs increase moderately when adding anchors, as shown in Table~\ref{tab:AppendFlops}. The FPS drops slightly mainly since multiple anchors incur inevitable more items processed in the time-costing NMS procedure. 

\section*{Appendix III. Visualization}
\AppendTabA
\AppendFigC
\textbf{Consistency.}
As shown in Figure 1 in the main paper, our proposed MuSu alleviates the inconsistency between the classification and regression head suffered from the FCOS detector. Here, we provide more visualizations about this, shown in Figure~\ref{fig:AppendMore}.

\textbf{Visualization of detection results.} We present the detection results of the MuSu model with $\#A=3$ and R-50 backbone in Figure~\ref{fig:det}. We can find that the MuSu-trained detector tends to relate the classification score to how well the predicted bounding box localizes (especially in multiple people cases).

\AppendFigE

\clearpage
{\small
\bibliographystyle{ieee_fullname}
\bibliography{egbib}
}

\end{document}